\documentclass[runningheads]{llncs}
\usepackage{marvosym}
\usepackage{multirow}
\usepackage{booktabs} 
\usepackage{array}
\usepackage{algorithm}
\usepackage{algpseudocode}
\usepackage{tabularx}
\usepackage{amsmath}

\newcommand{\Rmnum}[1]{\expandafter\@slowromancap\romannumeral #1@}
\usepackage[doipre={doi:~}]{uri}
\usepackage{circledsteps}
\usepackage{tikz}  
\newcommand{\CircledNumber}[1]{%
    \tikz[baseline=(char.base)]{  
        \node[shape=circle,draw,inner sep=1pt] (char) {\footnotesize #1};  
    }  
}  
\usepackage{graphicx}

\usepackage{fancyhdr} 
\usepackage{lipsum} 

\fancypagestyle{plain}{\fancyhead[L]{\textit{This paper has been accepted for publication in SecureComm 2025.}}}
\begin{document}
\title{Privacy-preserving Preselection for Face Identification Based on Packing}
%
%
\author{
Rundong Xin\inst{1,4} 
\and
Taotao Wang\inst{1} \and
Jin Wang\inst{2} \and
Chonghe Zhao\inst{3} \and
Jing Wang\inst{4,5}\textsuperscript{\Letter}
}

\authorrunning{This paper has been received by SecureComm 2025}
%
\institute{College of Electronics and Information Engineering, and the Provincial Key Laboratory of Intelligent Communication and Digital Society Governance, Shenzhen University, Shenzhen, China
\and
Sanya Institute of Hunan University of Science and Technology, Sanya, China
\and
School of Computer Science and Cyber Engineering, Guangzhou University,
Guangzhou, China
\and
School of Computer Science and Technology, Changsha University of Science and Technology, Changsha, China
\and
School of Civil and Environmental Engineering, Changsha University of Science and Technology, Changsha, China\\ \email{znwj\_cs@csust.edu.cn}
}

%
\maketitle              
\begin{abstract}
Face identification systems operating in the ciphertext domain have garnered significant attention due to increasing privacy concerns and the potential recovery of original facial data. However, as the size of ciphertext template libraries grows, the face retrieval process becomes progressively more time-intensive. To address this challenge, we propose a novel and efficient scheme for face retrieval in the ciphertext domain, termed Privacy-Preserving Preselection for Face Identification Based on Packing (PFIP). PFIP incorporates an innovative preselection mechanism to reduce computational overhead and a packing module to enhance the flexibility of biometric systems during the enrollment stage. Extensive experiments conducted on the LFW and CASIA datasets demonstrate that PFIP preserves the accuracy of the original face recognition model, achieving a 100\% hit rate while retrieving 1,000 ciphertext face templates within 300 milliseconds. Compared to existing approaches, PFIP achieves a nearly 50x improvement in retrieval efficiency.

\end{abstract}
\section{Introduction}

Modern facial recognition is widely used in the Internet of Things (IoT) and has great commercial value. \cite{bpl1,b18,bpl2}. However, recent studies have revealed that unprotected \cite{b17,b16} and even some protected \cite{b33} biometric templates can be exploited to reconstruct approximate facial images, thereby compromising biometric privacy \cite{badv}. This vulnerability has driven the development of various biometric template protection schemes \cite{b1}. To standardize biometric template protection, the ISO/IEC 24745 has established three core requirements for biometric authentication systems: unlinkability, irreversibility, and updatability \cite{b4}.

To comply with these requirements, fully homomorphic encryption (FHE) has demonstrated significant potential for application in the field of biometric systems \cite{bfhe1,ciphertext,bfh2}. Specifically, FHE provides robust protection for face features transmitted over communication channels and enables secure computation on the server side. Within this context, biometric systems typically operate in two distinct modes: \textit{Face Recognition \textnormal{(}$1:1$\textnormal{)}} and \textit{Face Authentication \textnormal{(}$1:n$\textnormal{)}} \cite{b11}. For \textit{Face Recognition \textnormal{(}$1:1$\textnormal{)}}, it performs the ciphertext-based Euclidean distance computation only once, and its recognition efficiency is significantly improved by proposing efficient homomorphic algorithms. For example, the homomorphic algorithms in \cite{b6} and \cite{b5} can complete the face recognition within 1 seconds. For \textit{Face Authentication \textnormal{(}$1:n$\textnormal{)}}, it requires matching a large number of face templates one by one so that the computational load grows exponentially. To mitigate this prohibitive computational demand, several workload reduction strategies \cite{b8,b7,b9} leveraging software optimization have been proposed, including packing, feature transformation, and preselection. 



Currently, feature transformation has made significant advancements, enabling the decomposition of features into 64 dimensions \cite{b2,b22}. However, packing \cite{b10,b2} and preselection \cite{b11,b7,b3,bbb} still face three primary challenges—one associated with packing and two with preselection. For packing, existing schemes require the collection of multiple face samples prior to encryption, which introduces additional latency. For preselection, the system must ensure both the security and accuracy of face recognition to prevent the omission of the target face from the candidate queue. Also, an uneven distribution within the candidate queue can lead to an unnecessarily large queue size, thereby increasing computational overhead and degrading overall system performance.


To overcome the above challenges for preselection and packing, we propose a scheme, namely Privacy-preserving \underline{P}reselection for \underline{F}ace \underline{I}dentification Based on \underline{P}acking (PFIP), which complies with the ISO/IEC 24745 security standards.

In particular, the following contributions to the aforementioned research problems are made.
\begin{itemize}

\item \textbf{Proposing a novel preselection scheme}: We devise an innovative template encapsulation structure and fully exploited the homomorphic computation mechanism. This integration enables our packing module to be seamlessly incorporated into existing biometric authentication frameworks, thereby significantly enhancing their flexibility and adaptability.

\item \textbf{Designing an optimized packing module}: We utilize a subset of facial features to construct the candidate queue. The simplicity of the feature extraction process obviates the need for complex face fusion computations and circumvents the necessity for additional models associated with soft biometric features. As a result, our approach achieves superior efficiency and accuracy.
\item \textbf{Achieving high performance}: We implement PFIP and set up detailed experiments for analysis. The experimental results show that the hit rate is 100\%, effectively protecting the accuracy of the original face recognition model. The computational load is reduced by 99.61\% compared with the exhaustive search. Compared to the leading preselection scheme, the PFIP speeds up nearly 30x.

\end{itemize}

\section{Related Work}
In this section, we review several biometric recognition works based on privacy preservation that are relevant to the PFIP. These schemes can be broadly classified into two categories: preselection and packing method. 

\subsection{Preselection Methods}
Facilitating the incorporation of feature packing with preselection mechanisms, our study has conducted a comprehensive investigation into privacy-preserving binning. Privacy-preserving binning is defined as the division of the complete face library into multiple groups \cite{b7}. The main types of binning schemes are face fusion and soft biometric attributes.

Face-fusion-based schemes combine multiple sets of face features into a single set of features. One advantage of this type of method is that P. Drozdowski employs ciphertext-based preselection \cite{b11,b3}. Nevertheless, the computational complexity of this approach is high and the operation is time-consuming. Although it is combined with the tree structure in the scheme \cite{b3} to accelerate the retrieval process which is still unbearable.

Privacy-Preserving Preselection for Protected Biometric Identification Using Public-Key Encryption With Keyword Search (PEKS) allows for the efficient retrieval of soft biometric attribute through the use of public key encryption with keyword search. However, soft biometrics-based preselection is also accompanied by two obvious drawbacks. One of the limitations of this approach is the uneven distribution of bins. The bins are more uneven, resulting in a greater aggregation of face attribute features for certain soft-biological keywords \cite{b10}. In contrast, in scenarios such as companies, campuses, and national ID systems, skin colour and race tend to be grouped together in a similar category. Secondly, there is a significant reliance on the accuracy of face attribute recognition. Furthermore, the accuracy of the face attribute recognition model is of great importance with regard to the recognition of attributes such as gender and age. In the event of an inaccurate model, a large candidate queue must be set, which is a risky approach. 
\subsection{Coefficient Packing}
Coefficient packing can broadly be defined as the operation of one instruction on multiple operands by encapsulating multiple eigenvalues in a single encrypted vector.

In 2017, the concept of coefficient packing is introduced for the first time, with the aim of accelerating the distance computation under the biometric ciphertext domain \cite{b13}. Nevertheless, the specific application of this approach to the feature matching computation for face recognition has not yet been implemented. 

In 2018, V. Naresh Boddeti applied coefficient packing to face recognition for the first time. They implemented the inner product operation during the Euclidean distance calculation process by utilizing the operation of the ciphertext rotation. The scheme can recognize a face (1:1) in 0.25 seconds \cite{b14}. Although it has been applied to face recognition systems with real-time response, applications involving face retrieval remains unattained. 

To have a great performance on multiple face, P. Bauspieß et al. built upon this foundation by packing the features of multiple faces into a ciphertext vector \cite{b2}, in 2022. Their experimental results demonstrated that the time consumed with the Cheon-Kim-Kim-Song (CKKS) algorithm \cite{b12} under a library of 1000 templates is approximately 3 seconds. This approach effectively reduces the number of computations. 

However, one disadvantage of this method is that the rotation operation for ciphertext rotation imposes a significant performance penalty. Furthermore, the method rotates one bit at a time, which results in a slow inner product operation and the inability to fully utilize the results of the history operation. To realize a fast innner product algorithm, the most efficient packing scheme utilises $2^{i}$ bits in a single move. For 128-bit features, a mere six rotations are employed \cite{b22}.

Based on the above schemes, we can see that both coefficient packing and preselection has powerful computing ability. Meanwhile, all the preselection methods are depending on much extra computation or extra models, limiting dramatically their usability and deployability in real-world scenarios. Therefore we design the PFIP with little extra computation and do not dependent on extra models. Furthermore, the intersection of preselection and coefficient packing greatly improves the efficiency of face recognition.
\section{Baseline System}
Prior to describing the PFIP, it is necessary to provide a comprehensive overview of the general face recognition architecture. At the meantime, to demonstrate the benefits of the PFIP, we compare it with the state-of-the-art packing-based solution. The specific process of the packing scheme is mentioned in this section.
\subsection{Architecture}
The architecture of a typical face recognition system consists of dual servers and an edge-side device \cite{b10}. Notably, the edge-side device does not handle any key material to prevent it from functioning as a two-factor authentication component. Instead, it is tasked with extracting biometric features and generating ciphertext templates using public keys. The dual servers include a computation server (CS) and an authentication server (AS). The CS is responsible for storing ciphertext biometrics and public keys, as well as performing computations during the preselection and fine matching stages. In contrast, the AS holds the private key and is tasked with decrypting and verifying the ciphertext. Given the critical role of the AS in the system, it is assumed to be honest but curious, ensuring that it cannot access the ciphertext templates stored in the CS.
\subsection{Face Retrieval Process}
The proposed solution leverages a state-of-the-art packing algorithm that fully exploits the coefficient slots, enabling the storage of multiple face templates within a single ciphertext. This approach allows for the computation of multiple face similarity values from a single ciphertext, thereby significantly enhancing the efficiency of the system.

\begin{algorithm}[t]
\caption{Optimized packing algorithm}
\renewcommand{\algorithmicrequire}{\textbf{Input:}}
\renewcommand{\algorithmicensure}{\textbf{Output:}}
\begin{algorithmic}[1]
\Require $f_d$, $c_p$, $p_k$
\Ensure $c_{distance}$
\State $f_p \gets f_d \times k$ \Comment{Stitch $k$ copies of the segmented features together.}
\State $P_p \gets f(f_p)$ \Comment{Conversion of Eigenvectors to Eigenpolynomials.}
\State $s_p \gets \text{Enc}(P_p, p_k)$
\State $ct \gets \text{Multiply}((s_p - c_p), (s_p - c_p))$
\State $ct_1 \gets ct$
\For{$i \gets 1$ to $\log m$}
    \State $ct \gets \text{Rotate}(ct, 2^i)$
    \State $ct_1 \gets \text{add}(ct, ct_1)$
    \State $ct \gets ct_1$
\EndFor
\State $c_{distance} \gets st$
\State \Return $c_{distance}$
\end{algorithmic}
\end{algorithm}

The generation of the key is accompanied by the setting of the degree of polynomial to $N$ and the dimension of a facial feature to $d$. The face retrieval implementation process is delineated in Algorithm 1. On the CS, each ciphertext $c_p$ contains $N/2d$ face templates. During face retrieval, the edge side device extracts face features through the face recognition model and makes $(N/2d)-1$ copies of them. Then these are glued together to form a packing vector $f_p$ of length $N/2$ dimensions. The $f_p$ is encrypted as $S_p$ and sent to the AS. On the AS, all ciphertexts $c_p$ in the ciphertext database are traversed and the Euclidean distance computation is performed through the fast packing algorithm. The final result of the ciphertext computation is then sent to AS, and the face information of the database corresponding to the feature is determined by threshold.
\section{Proposed System}
Our work is inspired by the contributions of P. Drozdowski \cite{b3}. In the face fusion scheme, if the target face is included among the multiple faces being fused, the similarity calculation result is highly likely to be smaller compared to other fused faces. However, face fusion introduces a certain degree of computational overhead. Meanwhile, we observe that lower-dimensional face features significantly accelerate the speed of face recognition. Therefore, decomposed features are employed in the preselection stage. 

In this section, we elaborate on the design of the packing module for flexible face recognition and the detailed workflow of the PFIP. The notation used throughout this section is summarized in Table~\ref{note}.
\begin{table}[t]
\caption{Notations used in this section.}
\centering
\setlength{\tabcolsep}{12pt} 
\begin{tabular}{ll} 
\toprule
Notation & description                        \\ \midrule
$f_d$        & The face feature of $d$ dimension    \\
$f_m$        & The decomposed feature of $m$ dimension \\
$f_p$        & The packing-based feature          \\
$P_p$        & The polynomial of $f_p$            \\
$S_p$        & The decrypt of $f_p$                \\
$S_d$        & The decrypt of $f_d$                \\
$D_{divided}$ & The database of divided feature    \\
$D_{complete}$ & The database of complete feature   \\
$N$             & Half of the Poly\_module\_degree   \\ \bottomrule
\end{tabular}
\label{note}
\end{table}
\subsection{Packing Module for Flexible Face Enrollment}
In the original packing scheme, it is necessary to wait for \( k \) faces to sufficiently fill the ciphertext coefficient slots \cite{b10}. This requirement necessitates sending the ciphertext to the database only after accumulating a certain number of face templates, thereby introducing latency in the registration process.

A packing module is introduced to enhance the flexibility of packing for face enrollment scenarios, thereby addressing the identified limitations. The following section elaborates on the operational dynamics of the designed packing module. The edge-side device maintains a counter tracking the number of face templates stored in the latest bin. To illustrate, consider a scenario where the current counter value is \( k \), the face feature dimension is \( m \), and the capacity of a bin is limited to \( \frac{N}{m} \) templates (where \( N \) is half of the polynomial module degree). This implies that the latest bin in the database contains \( k \) face templates. Upon receiving a new \( n \)-dimensional face feature, the edge-side device pads the feature with \( k \cdot n \) zeros at the front and \( \left(\frac{N}{m} - k - 1\right) \cdot n \) zeros at the back. The padded feature is then encrypted into a ciphertext and transmitted to the database. On the CS, the ciphertext undergoes a homomorphic addition with the latest bin stored on the CS. After successful registration of the face feature, the counter on the edge-side device increments to \( k + 1 \). If \( k + 1 \) reaches \( \frac{N}{m} \), indicating that the bin is fully occupied, the counter resets to 0, and the packing module completes the registration process more flexibly.
\subsection{Progress of PFIP}
\begin{figure}[t]
    \centering
    \includegraphics[width=1\linewidth]{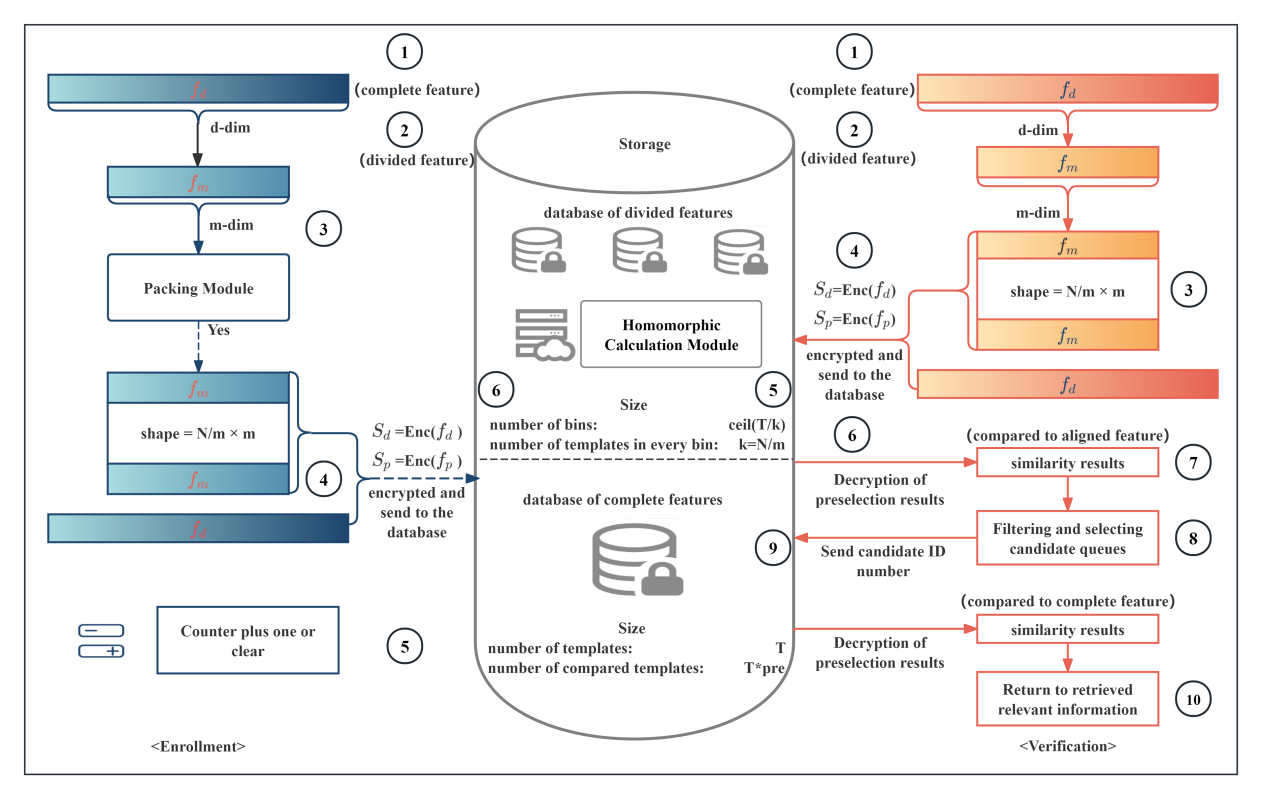}
    \caption{The workflow of the PFIP scheme.}
    \label{PFIP}
\end{figure}
The Fig. \ref{PFIP} illustrates the specific flow of the PFIP scheme in the context of registration and authentication. 

During the registration process:

\CircledNumber{1} On the edge-side, a face feature is extracted and normalization operations are then performed. The complete d-dimension feature is noted as $f_{d}$. 

\CircledNumber{2} The feature $f_{d}$ from one to m is extracted and represented by $f_{m}$. 

\CircledNumber{3} According to recorded number k in counter, the decomposed features $f_{m}$ is spliced $k\cdot n$ zeros in front of the feature and $(N/m-k-1)\cdot n$ zeros behind it, through packing module. And the result is denoted as $f_p$

\CircledNumber{4} The segmented ciphertext feature $S_{p}\leftarrow Enc(f_{p}, publickey)$ and the complete feature $S_{d}\leftarrow Enc(f_{d}, publickey)$ are uploaded to the CS.

\CircledNumber{5} The value of $k$ is adjusted to $k+1$ unless $k+1 = N/m$,, in which case $k$ is reset to 0.

\CircledNumber{6} The server side incorporates the ciphertext template $S_p$ into the ciphertext database $D_divided$ by employing once homomorphic addition.

During the verification process: 

\CircledNumber{1} and \CircledNumber{2} On the edge-side, the probe is executed the same process as that employed during the registration phase. 

\CircledNumber{3} The $f_{m}$ spliced with itself $[N/m]$ times. Thus, $f_{p}$ contains segmented features $[N/m]$ . 

\CircledNumber{4} The encrypted values of $S_{d}\leftarrow Enc(f_{d}, publickey)$ and $S_{p}\leftarrow Enc(f_{m}, publickey)$ are transmitted to the database server. 

\CircledNumber{5} On the CS, the cipher similarity calculation is initially performed on $S_{p}$ and bins from $D_{divided}$. 

\CircledNumber{6} The $result1\leftarrow \varSigma _{i=1}^{n}S_{pi}\cdot bin_i$ is then returned to the terminal. Subsequently, 

\CircledNumber{7} On the AS, the ciphertext is decrypted to $plainresult \leftarrow Dec(result1,\quad secretkey)$ 

\CircledNumber{8} The face candidate queue is screened out according to the set penetration rate. 

\CircledNumber{9} On the server side, the features that require further comparison are selected from $D_{complete}$, and the calculation is executed from $S_{d}$ and $S_D$ from $D_{complete}$. The $result2\leftarrow \varSigma _{i=1}^{n}( S_{di}-S_{Di})^2$ should be returned to the terminal. 

\CircledNumber{10} On the AS, the ciphertext is decrypted as plaintext. It is compared with the threshold and results are sent to edge-side.
\section{Experiment}
In section 5, we describe the PFIP implementation details. To reflect the performance, efficiency and security of the PFIP. We design experiments to analyze the hit rate, time-consuming and workload reduction. We also compare the PFIP with the state-of-the-art scheme and analyze its compliance with ISO/IEC 24745.
\subsection{Experimental Setup}
The evaluation and testing of the proposed scheme was conducted on an Intel(R) Core(TM) i7-13650HX @ 2.60 GHz processor with 32GB of RAM and a 64-bit Windows 11 operating system. The computations were executed on a single thread. For the sake of reproducibility, the specific settings of the experiment were listed as below. The implementation of the PFIP can be divided into two parts: face recognition model, homomorphic encryption and preselection parameter settings. 

(1) Face recognition model: For face recognition, the MTCNN was selected to extract the face in the image. For face recognition, we selected Facenet \footnote[1]{https://github.com/davidsandberg/facenet} and Insightface \footnote[2]{https://github.com/deepinsight/insightface}. Facenet \cite{b16} encodes faces as 128-dimensional face feature vectors, while Insightface \cite{b17} encodes faces as 512-dimensional. 

(2) Parameter of homomorphic encryption: For protecting face templates, we used the CKKS algorithm from SEAL library \cite{b21}. There were two ciphertext database: $D_{divided}$ and $D_{complete}$. For $D_{divided}$, the key parameters polynomial degree, coefficient modulus and scaling factor were set to 32768, \{40, 30, 40\} and $2^{30}$, respectively. For $D_{complete}$, the degree of polynomial, coefficient modulus and scaling factor were set to 8192, \{40, 30, 40\}, $2^{30}$, respectively. The scaling factor was set to $2^{30}$, which ensures the demand for high accuracy in face recognition applications. This guaranteed that the computation results would not affect the performance of face recognition after decryption.

(3) Parameter of preslection schemes: Through experiments on preselection, the optimal parameter settings were selected. The Facenet model had a decomposed feature dimension of 64 and a penetration rate of 0.02. The Insightface model had a decomposed feature dimension of 128 and a penetration rate of 0.05. Penetration rate indicates the proportion of candidate cohorts screened. That is, for a total number of 1000 faces, the size of the candidate queue for screening is 50.
\subsection{Analysis of Preselection Hit Rates}
In this analysis, our objective was to pre-screen data without compromising the performance of the original face recognition models. To achieve this, we conducted data preprocessing prior to experimentation. Specifically, we employed FaceNet and InsightFace to filter and retain data that could be accurately recognized by these models from the LFW (Labeled Faces in the Wild) and CASIA datasets. This approach ensured that the selected data maintained high recognition accuracy while minimizing the impact on the overall performance of the face recognition system.
\begin{table}[t]
\caption{The penetration in different scenarios under the PFIP. (HR is the abbreviation of Hit Rate.)}
\label{hit}
\centering
\begin{tabularx}{\textwidth}{l*{5}{X}} 
\toprule
\multirow{2}{*}{Feature extractor}    & \multirow{2}{*}{Datasets} & \multirow{2}{*}{$d$} & \multicolumn{3}{c}{Penetration at} \\ \cmidrule{4-6} 
                                      &                           &                      & 99\%HR    & 99.5\%HR    & 100\%HR   \\ \midrule
\multirow{8}{*}{Insightface(512-dim)} & \multirow{4}{*}{CASIA}    & 256                        & 0         & 0           & 0.01      \\
                                      &                           & 128                        & 0.005     & 0.01        & 0.04      \\
                                      &                           & 64                         & 0.005     & 0.07        & 0.25      \\
                                      &                           & 32                         & 0.05      & 0.27        & 0.52      \\ \cmidrule{2-6} 
                                      & \multirow{4}{*}{LFW}      & 256                        & 0         & 0.005       & 0.02      \\
                                      &                           & 128                        & 0.005     & 0.02        & 0.05      \\
                                      &                           & 64                         & 0.02      & 0.1         & 0.24      \\
                                      &                           & 32                         & 0.06      & 0.32        & 0.53      \\ \midrule
\multirow{6}{*}{Facenet(128-dim)}     & \multirow{3}{*}{CASIA}    & 64                         & 0.005     & 0.01        & 0.02      \\
                                      &                           & 32                         & 0.02      & 0.1         & 0.2       \\
                                      &                           & 16                         & 0.06      & 0.24        & 0.48      \\ \cmidrule{2-6} 
                                      & \multirow{3}{*}{LFW}      & 64                         & 0.005     & 0.01        & 0.02      \\
                                      &                           & 32                         & 0.02      & 0.13        & 0.25      \\
                                      &                           & 16                         & 0.07      & 0.3         & 0.5       \\ \bottomrule
\end{tabularx}
\end{table}
\begin{figure}[t]
    \centering
    \includegraphics[width=1\linewidth]{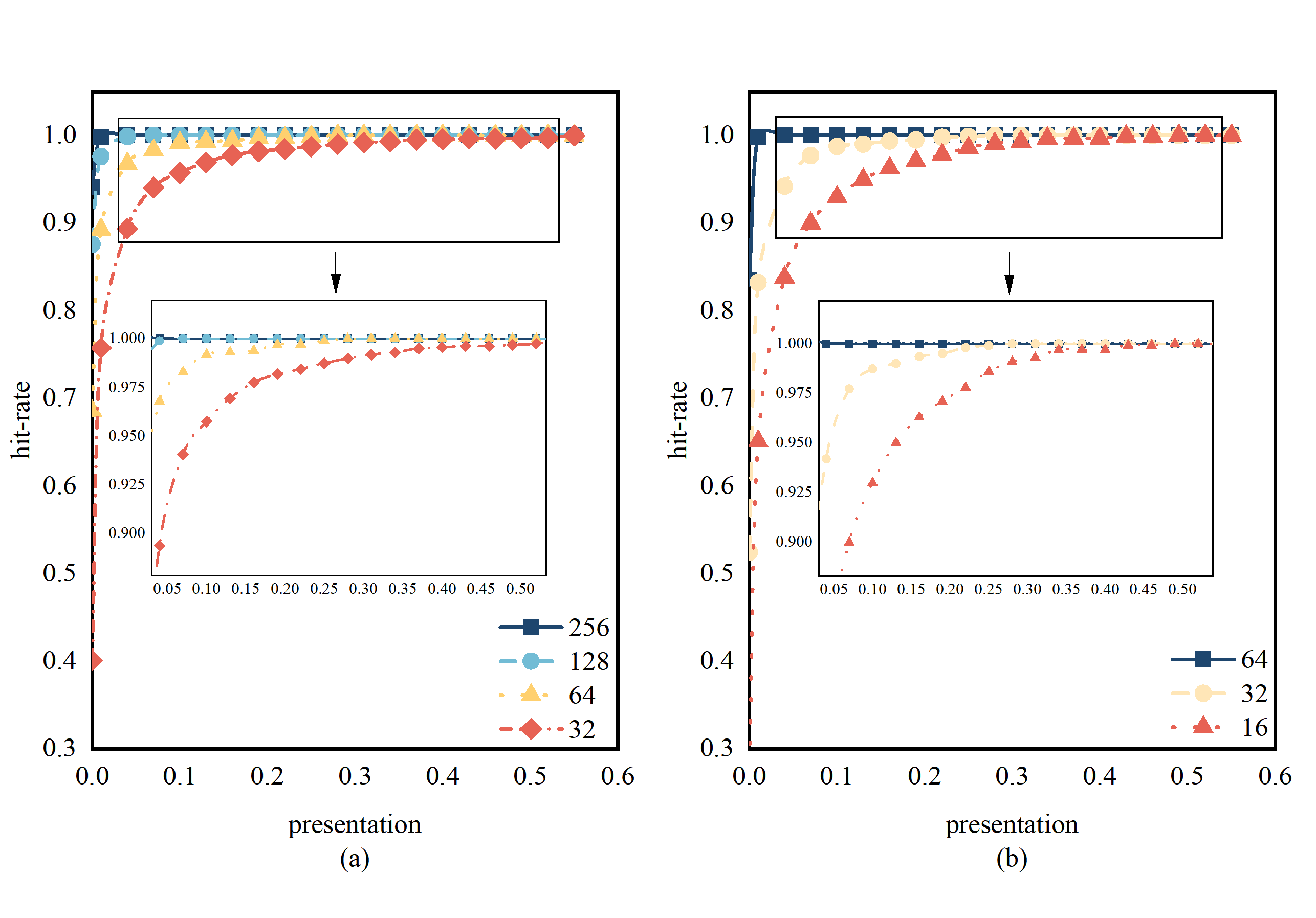}
    \caption{The hit-rate under differnet penetration.}
    \label{hitrate}
\end{figure}

For evaluating the performance of the preselection process, face images from the CASIA and LFW datasets are extracted according to the aforementioned criteria. Specifically, 3000 to 4000 subjects are randomly selected from the CASIA dataset, while 1000 to 1500 subjects are chosen from the LFW dataset. As shown in Table~\ref{hit}, the InsightFace model is tested with decomposed dimensions of 256, 128, 64, and 32, respectively. The candidate queues in both datasets are constructed to reflect the worst-case scenario of penetration corresponding to the target feature hit rate. Table~\ref{hit} demonstrates that the performance of the penetration rate under the same face recognition model is similar across different datasets. 

During the experimental recording of the worst-case penetration rate scenario, it is observed that the penetration rate stabilizes as the number of face templates increases to hundreds or thousands. As illustrated in Fig.~\ref{hitrate}, the penetration rate exhibits a trend where a higher hit rate corresponds to a larger candidate queue. Moreover, an increase in the dimensionality of the decomposed face features indicates a greater amount of face information, which is advantageous for narrowing down the candidate queue. The InsightFace model achieves a 100\% hit rate with 256- and 128-dimensional face features before the penetration rate reaches 0.1. For 64- and 32-dimension, the model also reaches a 90\% hit rate more quickly, although achieving a 100\% hit rate requires a larger subset of templates. Similarly, the FaceNet model's 128-dimension converge to 100\% more rapidly when decomposed into 64- and 32-dimension. In summary, if an accurate penetration rate is selected, the face recognition model can still achieve satisfactory performance.

\subsection{Performance Analysis of PFIP}
Analyzing the impact of PFIP on reducing computational load involves comparing it with the computational requirements of non-batch exhaustive searches. The reduction in computational load can be encapsulated by the following formula:

For $All=n\cdot d$, $d$ denotes the dimension of feature extraction and $n$ denotes the total number of face libraries. Thus, $All$ is the total computational load consumed by the exhaustive search. 

And the formula for $Real$ is given by $Real=(m\cdot n+\lceil p\cdot n\rceil \cdot d)/(N/d)$. In this formula, $m$ denotes the dimension of the extracted features, $p$ denotes the proportion of the candidate queue and $N$ denotes the half of Poly\_modulus\_degree. Thus, $Real$ is the actual computational load consumption after optimization. The proportion of the final computational load reduction can be expressed by Equation (1). The computational load is not affected by the total volume of the template library.
\begin{equation}
Workload=\frac{Real}{All}\approx \small{\frac{(m+p\cdot d)\cdot N}{d\cdot d}}
\end{equation}
\begin{table}[t]
\caption{Computational Workload Reduction and Time Consumption of Face Retrieval in Different Scenarios under the PFIP. (Pres. is the abbreviation of Preselection.)}
\label{perform}
\centering
\begin{tabular}{cccccccc}
\toprule
\multirow{3}{*}{$d$} & \multicolumn{6}{c}{Number of face templates}                                                                                                                                                                                                                                                                                                                                            & \multirow{3}{*}{\begin{tabular}[c]{@{}c@{}}Workload \\ reduction\end{tabular}} \\ \cmidrule{2-7}
                   & \multicolumn{2}{c}{1000}                                                                                                    & \multicolumn{2}{c}{3000}                                                                                                    & \multicolumn{2}{c}{5000}                                                                                                    &                                                                                \\ \cmidrule{2-7}
                   & \begin{tabular}[c]{@{}c@{}}Pres. \\ time(s)\end{tabular} & \begin{tabular}[c]{@{}c@{}}Fine matching \\ time(s)\end{tabular} & \begin{tabular}[c]{@{}c@{}}Pres. \\ time(s)\end{tabular} & \begin{tabular}[c]{@{}c@{}}Fine matching \\ time(s)\end{tabular} & \begin{tabular}[c]{@{}c@{}}Pres. \\ time(s)\end{tabular} & \begin{tabular}[c]{@{}c@{}}Fine matching \\ time(s)\end{tabular} &                                                                                \\ \midrule
256                & 0.64                                                     & 0.1                                                              & 1.88                                                     & 0.3                                                              & 3.16                                                     & 0.5                                                              & 1.625\%                                                                        \\
128                & 0.296                                                    & 0.25                                                             & 0.88                                                     & 0.75                                                             & 1.48                                                     & 1.25                                                             & 0.938\%                                                                        \\
64                 & 0.124                                                    & 1.3                                                              & 0.372                                                    & 3.9                                                              & 0.62                                                     & 6.5                                                              & 1.171\%                                                                        \\
32                 & 0.056                                                    & 2.65                                                             & 0.168                                                    & 7.95                                                             & 0.28                                                     & 13.25                                                            & 1.852\%                                                                        \\ \midrule
64                 & 0.124                                                    & 0.1                                                              & 0.372                                                    & 0.3                                                              & 0.62                                                     & 0.5                                                              & 0.406\%                                                                        \\
32                 & 0.056                                                    & 1.3                                                              & 0.168                                                    & 3.9                                                              & 0.28                                                     & 6.5                                                              & 0.391\%                                                                        \\
16                 & 0.024                                                    & 2.5                                                              & 0.072                                                    & 7.5                                                              & 0.12                                                     & 12.5                                                             & 0.512\%                                                                        \\ \bottomrule
\end{tabular}
\end{table}
Assessing retrieval time and computational load under varying penetration rate dimensions ensures a fair comparison, with evaluations conducted under scenarios maintaining a 100\% hit rate as the face template library expands. Such an approach safeguards the original model's performance from compromise.

For the InsightFace model, the penetration rates for decomposed features ranging from 256 to 64 dimensions are 0.02, 0.05, 0.25, and 0.53, respectively. For the FaceNet model, the corresponding rates for features ranging from 64 to 16 dimensions are 0.02, 0.25, and 0.50. As shown in Table~\ref{perform}, these rates provide a comprehensive analysis of the impact of different segmentation schemes on the computational workload and ciphertext operation overhead under the specified parameter settings for both models.

\begin{figure}[t]
        \centering
        \includegraphics[width=1\linewidth]{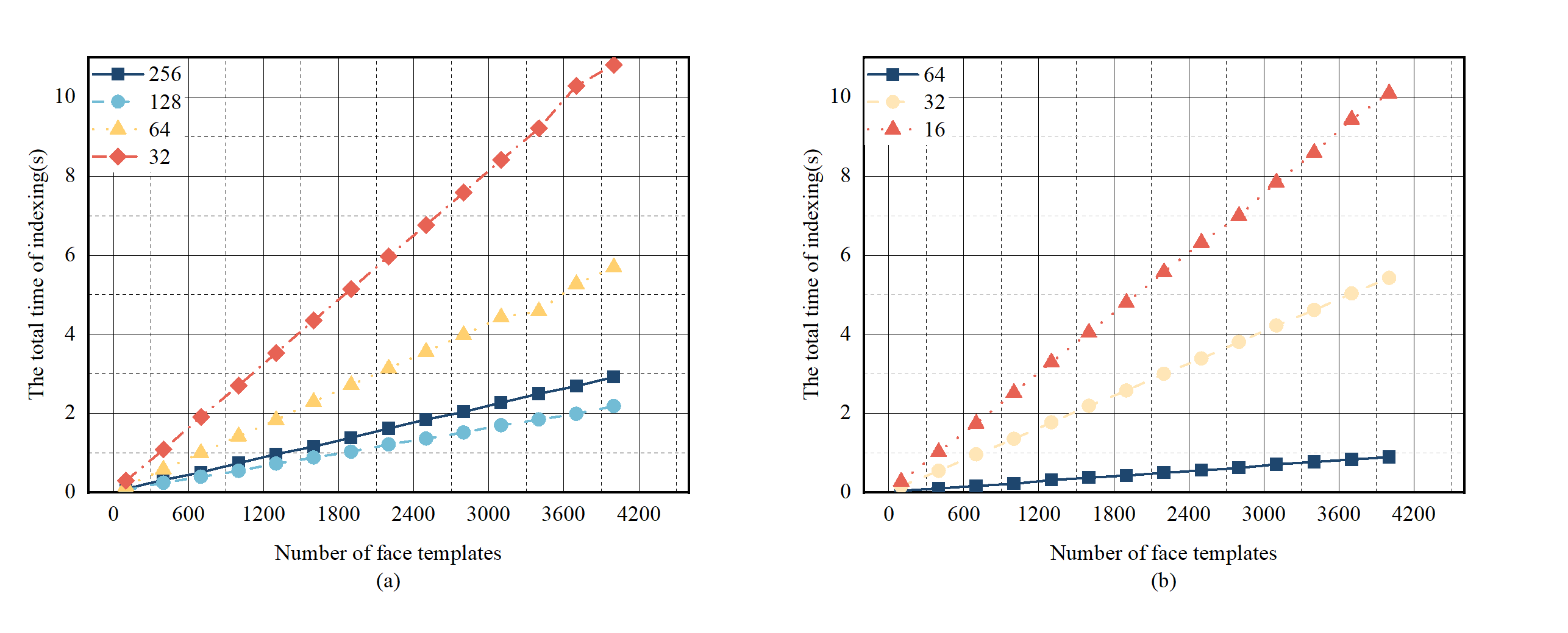}
        \caption{The total time of the PFIP scheme in different face recognition model. (a): Insightface model (512-dim). (b): Facenet model (128-dim).}
        \label{performance}
\end{figure}

In the preselection stage, smaller decomposed features allow more templates to be packed into a single ciphertext, thereby accelerating retrieval. However, this also results in a larger candidate queue, which significantly increases the time required for exact matching. As illustrated in Fig.~\ref{performance}, combining the total time consumed for ciphertext retrieval under different preselection parameters reveals that optimal retrieval performance is achieved with decomposed features of 128 dimensions for the InsightFace model and 64 dimensions for the FaceNet model. This approach reduces the computational workload by 93.8\% and 40.6\%, respectively, compared to the original exhaustive search.
\subsection{Performance Comparison}
\begin{table}[t]
\caption{Comparison of different schemes.}
    \label{compare}
    \centering
    \setlength{\tabcolsep}{10pt} 
\begin{tabular}{l>{\centering\arraybackslash}p{0.5cm}>{\raggedright\arraybackslash}p{2cm}>{\raggedright\arraybackslash}p{2cm}>{\raggedright\arraybackslash}p{2cm}}
\toprule
Scheme                    & $d$   & Ciphertexts retrieval time(s) & Total time(s) & Workload reduction \\ \midrule
\multirow{2}{*}{PEKS \cite{b10}} & 128 &  --\hspace{1cm}                     & 1.39          &   --      \\
                          & 512 &  --\hspace{1cm}                    & 19.25         &   --     \\ \midrule
\multirow{2}{*}{Baseline} & 128 & 0.29                         & 0.38          & 0.781\%            \\
                          & 512 & 1.373                        & 1.463         & 3.125\%            \\ \midrule
\multirow{2}{*}{PFIP}      & 128 & 0.224                        & \textbf{0.334}         & \textbf{0.398\%}            \\ 
                          & 512 & 0.546                        & \textbf{0.656}         & \textbf{0.938\%}            \\ \bottomrule
\end{tabular}
\end{table}
We provides a comprehensive record of the execution time of the ciphertext computation, the total search time, and the reduced computational load for a range of scenarios involving 1000 ciphertext template libraries and various face recognition models in Table \ref{compare}. It should be noted that the total time encompasses both the ciphertext encryption and decryption time, as well as the ciphertext computation execution time. A comparison with the baseline reveals that our scheme compresses 20\% of the original computation time by recognizing a face in 33 milliseconds on average for 128-dimensional features. However, for 512-dimensional features, the computation time is more than doubled. Concurrently, the computational overhead is reduced by almost 2 and 4 times, respectively, on the basis of the original load. In comparison with the current state-of-the-art preselection model, the CKKS algorithm can expedite the search of 1000 ciphertext template libraries by 1s and 15s for 128 and 512-dimensional face features, respectively. Our experiments demonstrate that decomposed features retain the preselection property. By directly encrypting the decomposed features, the proposed PFIP not only avoids extensive homomorphic computations but also eliminates the need for additional models that might affect performance, such as those used in PEKS \cite{b10}.
\begin{figure}[t]
    \centering
    \includegraphics[width=1\linewidth]{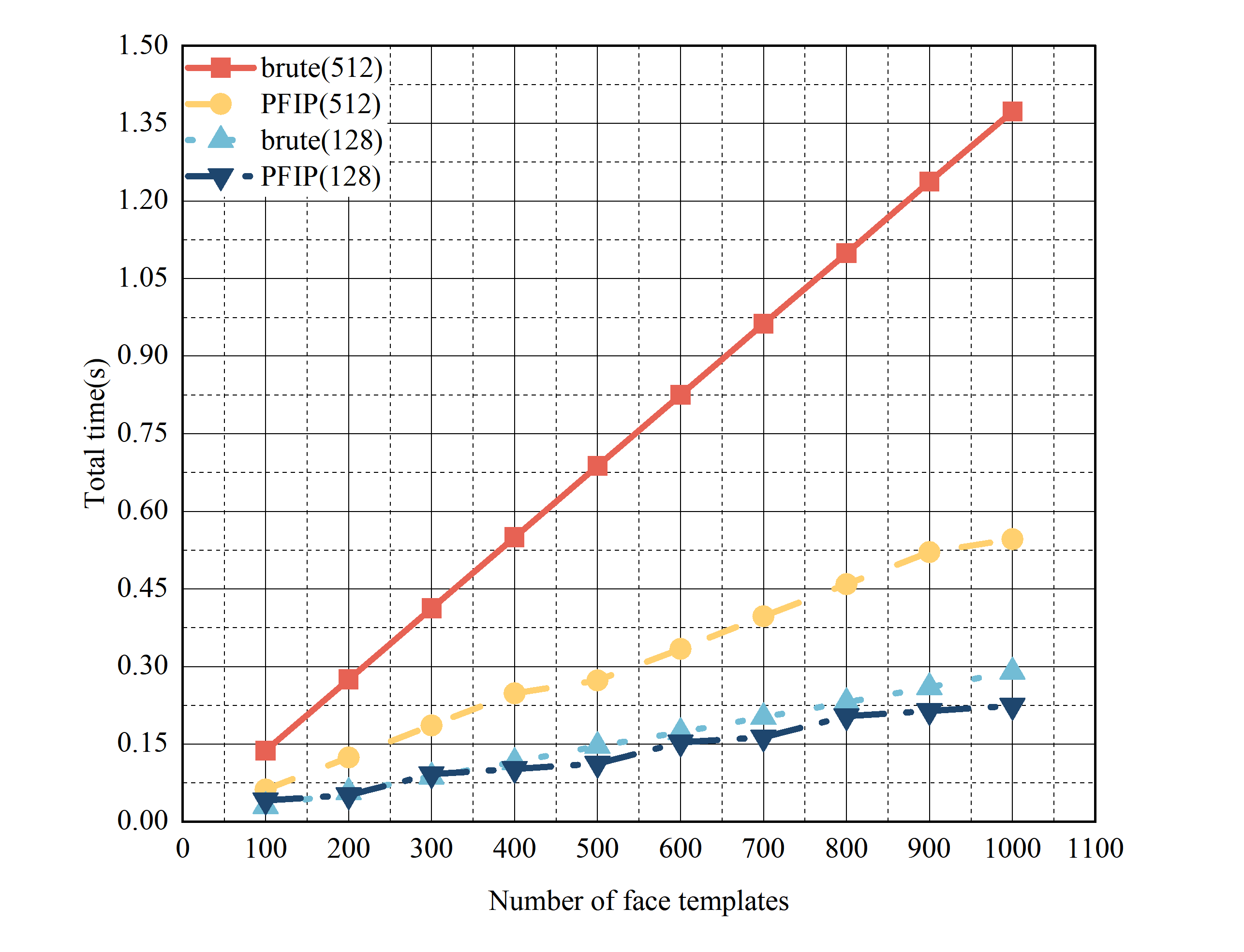}
    \caption{The consuming of time under different schemes.}
    \label{time-consuming}
\end{figure}
As demonstrated in Fig. \ref{time-consuming}, our scheme exhibits significant advantages in the field of face retrieval. When the face template library is limited, there is minimal variation in the number of packed ciphertexts. For instance, when 100 templates are selected for the facenet model, both the baseline and our method store one packed ciphertext. It is important to note that while preselection does not compromise the accuracy of the face recognition model, it does necessitate additional calculations in the candidate queue, thereby rendering its advantage less pronounced. However, as the template library expands, the number of stored packed ciphertexts experiences a substantial decline compared to the baseline. Consequently, our scheme exhibits linear growth at a significantly slower rate as the template library increases. Concurrently, the result can be extrapolated linearly based on the trend of the curve. When the template library reaches 1 million sheets, our scheme requires approximately 300 seconds.It is important to note that our evaluation is conducted under single thread; both the baseline and our method can be further enhanced in real-world applications through parallel computing.

\subsection{Security Analysis}
We evaluated the unlinkability, irreversibility, and updatability of the PFIP based on the ISO/IEC 24745 standard \cite{b4}. In addition, the proposed method achieves a balance between security and accuracy, ensuring robust privacy protection while exerting minimal impact on the precision of the results.

\textbf{Renewability:} When biometric templates are compromised, new and secure templates can be generated from the same source using public-key cryptography. This ensures that the system remains secure even if the original templates are leaked.

\textbf{Unlinkability:} The same subject should appear different across various applications. This is achieved through randomization techniques in the Fully Homomorphic Encryption (FHE) process, which prevent inferring any relationships between ciphertexts encrypted from the same feature.
\begin{table}[t]
\caption{Security level of CKKS with different secret key parameter settings.}
\label{security}
\centering
\setlength{\tabcolsep}{12pt} 
\begin{tabular}{p{2cm}p{3cm}p{2cm}}
\toprule
$N$                      & Security level & $logq$ \\ \midrule
\multirow{3}{*}{8192}  & 128            & 218  \\
                       & 192            & 152  \\
                       & 256            & 118  \\ \midrule
\multirow{3}{*}{32768} & 128            & 881  \\
                       & 192            & 611  \\
                       & 256            & 476  \\ \bottomrule
\end{tabular}
\end{table}

\textbf{Irreversibility:} The plaintext cannot be derived from the protected templates. The CKKS algorithm, based on approximate homomorphic encryption, leverages the Ring Learning with Errors (RLWE) problem, which poses a significant challenge due to its computational complexity. Defined within a specific algebraic structure, the RLWE problem is used to construct secure homomorphic encryption schemes against quantum adversaries. As shown in Table~\ref{security}, the security level depends on the configuration of various parameters. The polynomial modulus degree is denoted by \( N \), and the coefficient modulus bit length is denoted by $logq$. Based on the experimental parameters, when \( N \) is set to 8192 or 32768, the coefficient modulus bit length is \( 40 + 30 + 40 = 110 \), which is less than the bit length required for a security level of 256. Consequently, the security level of the biometric template is guaranteed to be higher than 256 bits. This level of security is deemed sufficient for encrypting face templates.

\textbf{Performance:} To optimize the homomorphic encryption parameters for the PFIP, we carefully designed the polynomial degree settings as {40,30,40}. This configuration ensures high precision in the decryption process within the PFIP, as the Euclidean distance computation involves only a single multiplication operation. Additionally, the scaling factor was set to $2^{30}$, resulting in a precision loss of approximately $10^{-5}$ to $10^{-6}$ compared to plaintext computation. This level of precision loss is negligible and does not impact the accuracy of face recognition. Therefore, the PFIP achieves high-precision and high-security face authentication.
\section{Conclusion}
The article showed that PFIP have the best performance and high security among the present schemes. Utlizing the decomposed features, PFIP avoids much homomorphic computation and preserves the performance of the original face recognition model. Through the experiments, we can see that the system, combined with packing and preselection, has significant reduction about 99.6\% in computational load. PFIP retrieves 1000 faces in 300 and 400 ms, which speed up nearly 50x. The preselection scheme has been demonstrated to possess significant research value in accelerating the face retrieval problem, and the concept underlying the scheme can be applied to other biometric authentication methods.
\\
\\
\textbf{Acknowledgements} This research is supported by the National Natural Science Foundation of China (Grant No. 2023JJ50331) and Hunan Provincial Education Department Scientific Research Project of China (Grant No. 23C0123). Jing Wang is the corresponding author. This research is supported by Research Center for FinTech and Digital-Intelligent Management, Shenzhen University.
\end{document}